\documentclass{article}

    \usepackage[preprint]{main}

\usepackage[utf8]{inputenc} 
\usepackage[T1]{fontenc}    
\usepackage{hyperref}       
\usepackage{url}            
\usepackage{booktabs}       
\usepackage{amsfonts}       
\usepackage{nicefrac}       
\usepackage{microtype}      
\usepackage{xcolor}         
\usepackage[pdftex]{graphicx}
\usepackage{amsmath}
\usepackage{multirow}
\usepackage{array}
\usepackage{float}
\usepackage{enumitem}
\usepackage{titlesec}

\titlespacing*{\section}{0pt}{*1.5}{*0}

\titlespacing*{\subsection}{0pt}{*1}{*0}

\title{How Does Diverse Interpretability of Textual Prompts Impact Medical Vision-Language Zero-Shot Tasks?}

\author{%
 Sicheng Wang \\
 Department of Earth Science and Engineering\\
 Imperial College London\\
 Exhibition Rd, South Kensington, London SW7 2AZ \\
 \texttt{tony.wang23@imperial.ac.uk} \\
 \And
 Che Liu\thanks{Corresponding author: \texttt{che.liu21@imperial.ac.uk}} \\
 Department of Earth Science and Engineering\\
 Imperial College London\\
 Exhibition Rd, South Kensington, London SW7 2AZ \\
 \texttt{che.liu21@imperial.ac.uk} \\
 \And
 Rossella Arcucci \\
 Department of Earth Science and Engineering\\
 Imperial College London\\
 Exhibition Rd, South Kensington, London SW7 2AZ \\
 \texttt{r.arcucci@imperial.ac.uk} \\
}

\begin{document}

\maketitle

\begin{abstract}
    Recent advancements in medical vision-language pre-training (MedVLP) have significantly enhanced zero-shot medical vision tasks such as image classification by leveraging large-scale medical image-text pair pre-training. However, the performance of these tasks can be heavily influenced by the variability in textual prompts describing the categories, necessitating robustness in MedVLP models to diverse prompt styles. Yet, this sensitivity remains underexplored. In this work, we are the first to systematically assess the sensitivity of three widely-used MedVLP methods to a variety of prompts across 15 different diseases. To achieve this, we designed six unique prompt styles to mirror real clinical scenarios, which were subsequently ranked by interpretability. Our findings indicate that all MedVLP models evaluated show unstable performance across different prompt styles, suggesting a lack of robustness. Additionally, the models' performance varied with increasing prompt interpretability, revealing difficulties in comprehending complex medical concepts. This study underscores the need for further development in MedVLP methodologies to enhance their robustness to diverse zero-shot prompts.
\end{abstract}

\section{Introduction}
\label{sec:introduction}
Medical Vision Language Pre-training (MedVLP) is a rapidly developing topic within the machine learning community\cite{liu2023t3d,chen2023generative,liu2023m,liu2024imitate,wan2024med,liu2023g2d,liu2023utilizing,qin2024freeze}. For downstream tasks, models pre-trained with MedVLP is being applied to zero-shot diagnosis tasks, taking only image input and a textual prompt to describe the category name \cite{BioViL, BioViL-T, ConVIRT, GLoRIA, KAD, MAVL, MedKLIP}, thereby diminishing the data requirements and also enabling generalisation to open-set tasks. Recent Medical VLP(MedVLP) models, such as BioViL\cite{BioViL}, MedKLIP\cite{MedKLIP}, and KAD\cite{KAD}, have achieved superior performance in zero-shot diagnosis for chest X-ray (CXR) images with diverse and advanced pre-training techniques.

\begin{figure}
  \centering
  \includegraphics[width=0.8\linewidth]{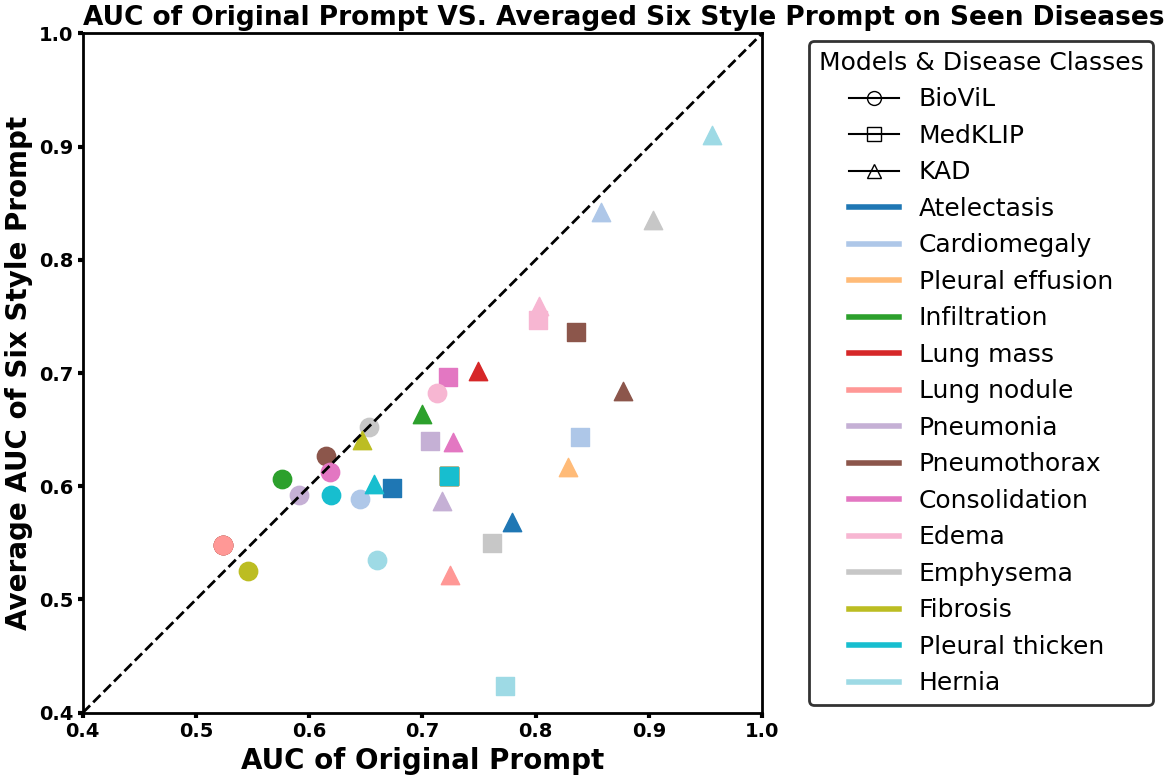}
  \caption{Comparison of original prompt and six style prompts' zero-shot image classification performance on seen disease classes of BioViL\cite{BioViL}, MedKLIP\cite{MedKLIP} and KAD\cite{KAD}. The X-axis shows the AUC performance of the original prompt, and the Y-axis shows the macro average of AUC performance of six style prompts. The dashed line shows the ideal scenario, where the model shows consistent performance on seen classes regardless of the prompt style.}
  \label{fig1}
\end{figure}

Our research identifies a critical limitation in MedVLP models: their high sensitivity to variations in textual prompts within zero-shot classification tasks. These models experience significant performance degrading when using diverse styles of textual prompts, as depicted in Figure \ref{fig1}. Ideally, a MedVLP model should provide consistent conclusions across disease classes, regardless of the prompt style, whether it uses simplified disease names or detailed CXR descriptions. This is crucial since clinicians can recognize the disease through various descriptions. Additionally, for disease classes not seen during pre-training, the MedVLP model should benefit from detailed, highly interpretable text prompts that provide comprehensive descriptions.

In our study, we utilise the large language model (LLM) GPT-4o \cite{GPT4o} to construct six different styles of text prompts for various diseases, each ranked by their interpretability. We evaluated three mainstream MedVLP models—BioViL\cite{BioViL}, MedKLIP\cite{MedKLIP}, and KAD\cite{KAD}—using these six styles of prompts on three publicly available benchmark test datasets: ChestX-ray14 \cite{ChestXray14}, CheXpert \cite{CheXpert}, and COVIDx CXR-4 \cite{COVIDxCXR4}. Quantitative evaluation revealed that when using prompts with styles different from those used in the original pre-training, the models' performance decreased by an average of 10.17\% in AUC score across all models, even when the diseases in the test set were already present in their pre-training datasets.

In addition to benchmarking MedVLP models with diverse textual prompts on zero-shot classification tasks, we analyse the varying degrees of performance degradation across different prompt styles. Based on these observations, we proposed a suggested retraining recipe for MedVLP models. This recipe is intended to help the community design robust MedVLP models that can effectively handle diverse textual prompts.

\section{Related Work}
\label{sec:related_work}
\subsection{General Vision Language Pre-training}
Vision-Language Pre-training (VLP) learns cross-modal representations from large-scale paired image-text data for various downstream tasks. Recent studies, such as CLIP \cite{CLIP}, ALIGN \cite{ALIGN}, ALBEF \cite{ALBEF}, and LiT \cite{LiT}, use contrastive learning on extensive multimodal datasets, scaling up both data and model sizes to enhance vision-language representation. Alternatively, approaches like BeiT3 \cite{BeiT3}, SLIP \cite{SLIP}, and A-FLIP \cite{A-FLIP} focus on making VLP more cost-effective by reducing model size and data requirements while maintaining high performance. In the medical domain, VLP applies to tasks like radiology report generation, disease diagnosis, and clinical decision-making. However, due to the fine-grained nature of these tasks and the need for clinical expertise, medical VLP (MedVLP) remains a significant challenge for ongoing research.

\subsection{Medical Zero-shot Classification Task}
By leveraging large, diverse datasets like radiology reports paired with CXR images \cite{MIMIC-CXR}, recent MedVLP models can identify diseases directly, without fine-tuning, by using their learned visual and textual representations to compute the similarity between input images and textual prompts. For instance, BioViL \cite{BioViL} redesigns vision-language models for better alignment with clinical texts, while BioViL-T \cite{BioViL-T} incorporates temporal data for enhanced zero-shot capabilities. ConVIRT \cite{ConVIRT} employs bidirectional contrastive learning to align medical images and text, and GLoRIA \cite{GLoRIA} uses an attention-based framework to learn global and local representations. MedKLIP \cite{MedKLIP} integrates external medical knowledge to improve zero-shot classification and grounding. MAVL \cite{MAVL} uses dual-head transformers in a multi-aspect description framework to enhance disease recognition. KAD \cite{KAD} leverages Unified Medical Language System (UMLS) knowledge graphs within a query-based transformer architecture for superior zero-shot performance in CXR diagnosis. Our study aims to thoroughly investigate the sensitivity of mainstream MedVLP models to text prompts, providing insights into their robustness and generalisation capabilities.

\subsection{Prompt Engineering for Zero-shot Task}
Recent studies \cite{improve1,improve2,improve3,improve4} aim to enhance general VLP models' zero-shot performance without costly retraining by using detailed, informative text prompts during inference. One such study, Xplainer \cite{Xplainer}, seeks to improve medical zero-shot diagnosis using a similar approach. However, Xplainer adopts a classification-by-description method and focuses on enhancing the explainability of zero-shot diagnosis, without fully exploring the models' adaptability to various prompt styles or thoroughly evaluating their sensitivity. Our research systematically investigates the performance of mainstream MedVLP models across diverse prompt styles with varying levels of interpretability.

\section{Methods}
\label{sec:methods}
\subsection{Overview}
In this study, we aim to evaluate the sensitivity of MedVLP methods to different textual prompts. Specifically, we focus on three mainstream MedVLP models: BioViL\cite{BioViL}, MedKLIP\cite{MedKLIP}, and KAD\cite{KAD}, which have demonstrated strong performance in zero-shot classification of CXR images in their original studies. In this section, we first provide an overview of the selected MedVLP methods, followed by the design of diverse prompt styles. These prompts are then ranked by interpretability using LLMs to further evaluate how the MedVLP models are affected by the level of prompt interpretability.

\subsection{Preliminary}
\begin{figure}[H]
  \centering
  \includegraphics[width=\linewidth]{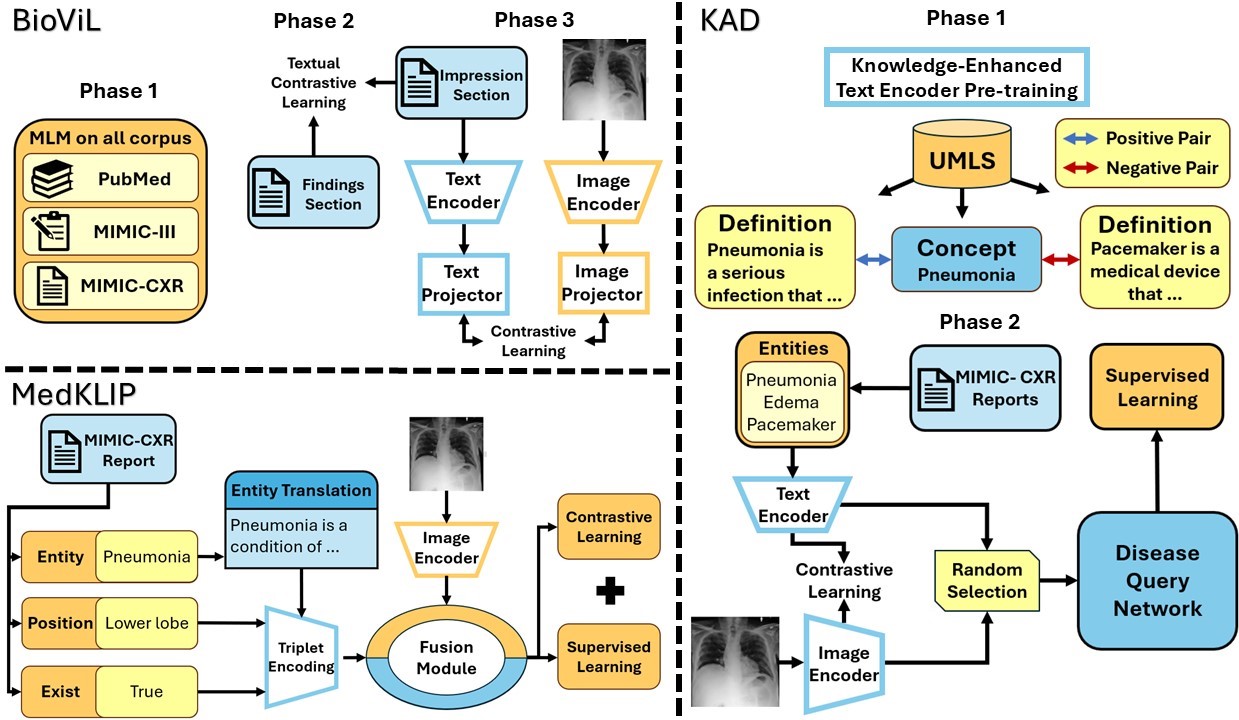}
  \caption{Framework of Three Mainstream MedVLP Models.
   \textbf{BioViL}: Phase 1 conducts a Masked Language Modelling (MLM) on a diverse corpus, including PubMed abstracts \cite{PubMed}, MIMIC-III clinical notes \cite{MIMIC-III}, and MIMIC-CXR radiology reports \cite{MIMIC-CXR}. Phase 2 involves textual contrastive learning between the Findings section and the Impression section of MIMIC-CXR reports. Phase 3 projects encoded image and text representations into a global space, then applies contrastive learning between them. 
\textbf{MedKLIP}: Pre-training involves extracting entity, position, and existence triplets from MIMIC-CXR reports. The model then translates simple entities into detailed descriptions and feeds these triplets into the fusion module together with encoded X-ray images. Lastly, it applies contrastive learning between image and text representations, and supervised learning based on the prediction results. 
\textbf{KAD}: Phase 1 pre-trains the knowledge-enhanced text encoder by applying contrastive learning between definition and concept pairs extracted from the Unified Medical Language System (UMLS) knowledge graph. Phase 2 applies combined contrastive learning between encoded entities and images and supervised learning on the disease query network by randomly selecting encoded entity and image pairs. 
  }
  \label{fig2}
\end{figure}

In this section, we introduce the three mainstream MedVLP methods utilised in our experiments to investigate the impact of diverse prompts on zero-shot CXR classification tasks. Additionally, we present the original prompt styles used in their respective studies
Notably, we did not retrain or re-implement their methods, as our focus is solely on zero-shot inference with diverse prompts. Therefore, we adopted the official code and pre-trained weights from the GitHub repositories provided by the original authors. The three MedVLP methods frameworks are shown in Figure \ref{fig2}.

\begin{itemize}[topsep=-3pt]
    \setlength{\itemsep}{1pt} 
    \setlength{\parskip}{1pt} 
    \setlength{\parsep}{1pt}  
    \item \textbf{BioViL\cite{BioViL}}: BioViL is one of the first studies to introduce a CXR domain-specific VLP model. BioViL employs advanced text augmentation, regularisation techniques, and multiple pre-training strategies, resulting in significant improvements in both image and text model performance across various medical benchmarks. The text inputs used in pre-training primarily consist of sentences from the Impression and Findings sections of MIMIC-CXR radiology reports (e.g., 'Specifically, no evidence of edema.' and 'There is no focal consolidation, pleural effusion, or pneumothorax.').
    \item \textbf{MedKLIP\cite{MedKLIP}}: MedKLIP utilises a unique triplet extraction module to simplify radiology reports into structured triplets. It focuses on integrating domain-specific knowledge into vision-language pre-training and introduces an entity translation module that leverages a medical knowledge base to translate simple entities into informative descriptions. The text inputs used in pre-training are descriptive sentences (e.g., 'Cardiomegaly, sometimes referred to as megacardia or megalocardia, is a medical condition in which the heart is enlarged.').
    \item \textbf{KAD\cite{KAD}}: KAD leverages an external knowledge graph \cite{KAD} to enhance auto-diagnosis for CXR. Additionally, KAD includes a transformer-based Disease Query Network (DQN) that uses disease names as queries for flexible zero-shot evaluations. The text inputs used in pre-training are simple entity names (e.g., 'pacemaker,' 'nodule,' 'pneumonia').
\end{itemize}

We also considered other notable studies in zero-shot diagnosis. For example, BioViL-T \cite{BioViL-T}, an enhancement of the original BioViL \cite{BioViL} architecture, performs well with data containing temporal information. However, it shows minimal improvements on standard datasets, making it less relevant for our study. MAVL \cite{MAVL} achieves impressive results with multi-aspect disease descriptions but requires specific prompt formats during inference, reducing its flexibility and making it incompatible with our approach.

\subsection{Design of Diverse Prompts}
In this section, we describe the pipeline as depicted in Figure \ref{fig3}. We constructed text prompts with varying levels of interpretability and styles to test the generalisation capabilities and sensitivity of the models. Specifically, we created six distinct prompt styles for each disease class:

\begin{figure}[H]
  \centering
  \includegraphics[width=0.8\linewidth]{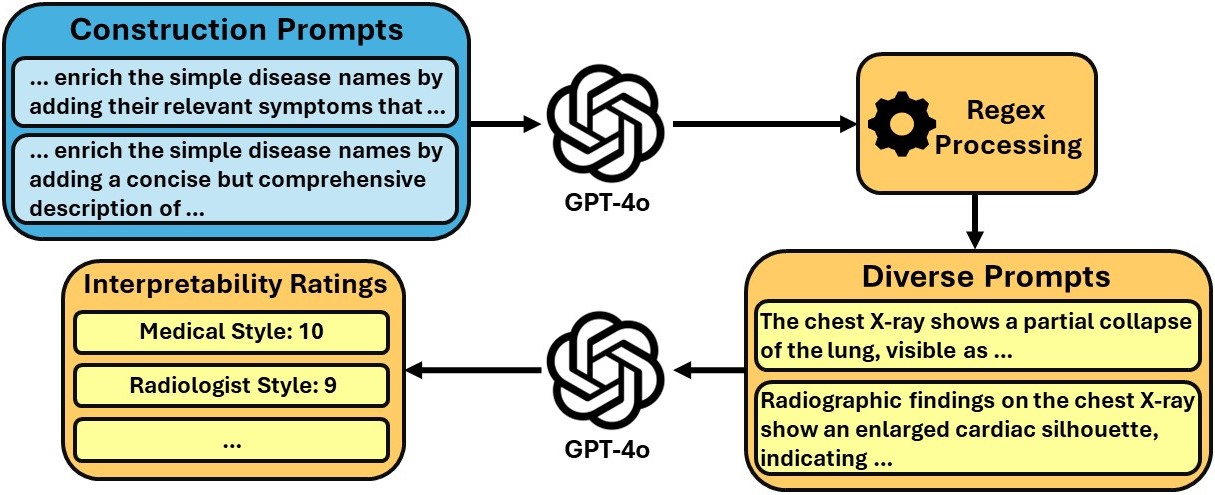}
  \caption{Pipeline of diverse prompt generation and interpretability score rating.}
  \label{fig3}
\end{figure}

\begin{itemize}[topsep=-3pt]
    \setlength{\itemsep}{1pt} 
    \setlength{\parskip}{1pt} 
    \setlength{\parsep}{1pt}  
    \item \textbf{Disease Name}: This style provides a concise, specific name for the condition without additional details. \newline
    \textbf{Example}: Atelectasis \newline
    \textbf{Interpretability Score}: \textbf{3}
    \item \textbf{Disease Symptom}: This style combines the disease name with key symptoms or radiographic findings associated with the condition. It focuses on the most prominent clinical and imaging characteristics. \newline
    \textbf{Example}: lung opacity, volume loss, mediastinal shift \newline
    \textbf{Interpretability Score}: \textbf{7}
    \item \textbf{Disease Attribute}: This style gives a detailed breakdown of the disease's features, including the appearance of borders, presence of fluid, location, opacity, patterns, shape, and texture. It provides a comprehensive description that helps in visualising the condition on imaging. \newline
    \textbf{Example}: border: clear but may show a shift of structures towards the collapsed area. fluid: no fluid accumulation typically present. location: localised to a segment, lobe, or entire lung. opacity: increased opacity in the affected area, appearing whiter than normal. other: displacement of structures like the diaphragm and trachea towards the affected side. patterns: linear or wedge-shaped opacity, with volume loss. shape: varies based on extent; can be a triangular or band-like appearance. texture: homogeneous texture within the collapsed region. \newline
    \textbf{Interpretability Score}: \textbf{9}
    \item \textbf{Disease Description(Plain English)}: This style describes the disease in straightforward, easy-to-understand language, focusing on how it would appear on a CXR imaging, and what that implies. It's intended for a general audience or non-specialists. \newline
    \textbf{Example}: The chest X-ray shows a partial collapse of the lung, visible as a white area where the lung tissue has lost its air content. This can appear as a dense, triangular shape pointing towards the hilum, and the affected area may cause a shift in nearby structures, such as the heart or trachea, towards the side of the collapse. \newline
    \textbf{Interpretability Score}: \textbf{8}
    \item \textbf{Disease Description(Radiologist Style)}: This style is used by radiologists and focuses on the interpretation of imaging findings. It provides a brief summary of what is seen on the X-ray or scan, and is intended for use in radiology reports or discussions among specialists. \newline
    \textbf{Example}: Radiographic findings on the chest X-ray demonstrate a partial collapse of the lung, often seen as increased density in the affected area. \newline
    \textbf{Interpretability Score}: \textbf{9}
    \item \textbf{Disease Description(Medical Style)}: This style provides a summary of the disease with clinical terms, often used in medical reports or documentation. It describes the imaging findings and their implications in a precise, formal manner, intended for healthcare professionals. \newline
    \textbf{Example}: Imaging reveals a collapse of lung tissue in the left lower lobe with volume loss and mediastinal shift towards the affected side. Bronchial obstruction is evident. \newline
    \textbf{Interpretability Score}: \textbf{10}
\end{itemize}

The GPT-4o prompts used to construct the diverse prompts are included in full as supplementary tables \ref{tab10}, \ref{tab11}, \ref{tab12}, \ref{tab13}.

\subsection{Prompt Interpretability Rating}
We again leveraged GPT-4o to rate the six prompt styles based on interpretability, using a scale from 1 to 10. A score of 10 represents the most interpretable and informative prompt for diagnosing the disease, while a score of 1 represents the opposite. This rating system helped us systematically assess the trend of the models' performance with different input prompt styles and interpretability levels. The GPT-4o prompt used to rate the interpretability of diverse prompts are provided as supplementary table \ref{tab14}.

\section{Experimental Setting}
\label{sec:experimental_setting}
\subsection{Datasets}
The three mainstream MedVLP models tested in this study, BioViL \cite{BioViL}, MedKLIP \cite{MedKLIP}, and KAD \cite{KAD}, primarily utilised the MIMIC-CXR \cite{MIMIC-CXR} dataset in their pre-training procedures. MIMIC-CXR is a publicly available dataset comprising 227,835 radiographic studies from 65,379 patients. Each study includes a corresponding CXR image and a free-text radiology report. Notably, since the MIMIC-CXR dataset was released prior to the COVID-19 pandemic, it does not contain any COVID-19 related cases.

To evaluate the generalisation capabilities of these models, we utilised three publicly available and widely-used datasets: ChestX-ray14\cite{ChestXray14}, CheXpert\cite{CheXpert}, and COVIDx CXR-4\cite{COVIDxCXR4}.

\begin{itemize}[topsep=-3pt]
    \setlength{\itemsep}{1pt} 
    \setlength{\parskip}{1pt} 
    \setlength{\parsep}{1pt}  
    \item \textbf{ChestX-ray14\cite{ChestXray14}} consists of 112,120 CXR images across 14 disease classes: Atelectasis, Cardiomegaly, Effusion, Infiltration, Mass, Nodule, Pneumonia, Pneumothorax, Consolidation, Edema, Emphysema, Fibrosis, Pleural Thickening, and Hernia. All 14 diseases have corresponding samples appearing in the MIMIC-CXR dataset. Our tests strictly followed the official train-test split, using a test set that includes 25,597 chest X-ray samples.
    
    \item \textbf{CheXpert\cite{CheXpert}} contains 224,316 CXR images. We used the official test set, which includes 500 CXR images annotated by radiologists. Following the original paper, our study focuses on the evaluation of 5 observations on the official test set: Atelectasis, Cardiomegaly, Consolidation, Edema and Pleural Effusion. All these classes have corresponding samples appearing in the MIMIC-CXR dataset. 
    
    \item \textbf{COVIDx CXR-4\cite{COVIDxCXR4}} is a major expansion of the dataset series COVIDx CXR-4. It includes 84,818 CXR images from 45,342 patients. The dataset has two classes: COVID-19 positive and COVID-19 negative. In this study, we used the official test set, which is perfectly class-balanced with 4,241 images in each category, totalling 8,482 images.
\end{itemize}

\subsection{Implementation}
To ensure fair comparison, all experiments were conducted on the same software environment and same device with RTX 3070 Mobile GPU. Before testing with our six prompt styles, we first evaluated each model on its baseline prompt style. The baseline style refers to the original prompt styles used in the respective studies: BioViL used "Findings suggesting + {disease name}", MedKLIP employed short disease descriptions, and KAD used the disease names alone. Performance on the baseline prompt serves as a benchmark, against which we compare the performance on other prompt styles to assess the models' true generalisation capabilities.

We then replaced the original prompts with the six different styles we constructed. All models used in the experiments were the original versions provided by the respective studies, without any further fine-tuning, to maintain a zero-shot setting. For KAD, which offers three different image encoder sizes (224px, 512px, and 1024px), we report results tested with the 512px-size encoder, as performance trends were similar across all sizes.

We used the image pre-processing methods described in the respective studies' original papers: For BioViL, we resized images to 512px and applied a 480px centre crop; For MedKLIP, we resized images to 224px and normalised them using global mean and standard deviation; For KAD, we resized images to 512px and normalised them using global mean and standard deviation. Similarly, text pre-processing was also performed according to the methods outlined in the original studies.

\section{Results and Analysis}
\label{sec:results_and_analysis}
We categorise the disease classes in our datasets into two groups: seen classes and unseen classes. 
\begin{itemize}
    \setlength{\itemsep}{1pt} 
    \setlength{\parskip}{1pt} 
    \setlength{\parsep}{1pt}  
    \item Seen classes are those that appear in the MIMIC-CXR dataset, which are used during the pre-training process of all three mainstream MedVLP models we focus on in this study.  This category includes 14 disease classes: Atelectasis, Cardiomegaly, Pleural effusion, Infiltration, Lung mass, Lung nodule, Pneumonia, Pneumothorax, Consolidation, Edema, Emphysema, Fibrosis, Pleural thicken and Hernia. These disease classes come from two datasets, \textbf{ChestX-ray14} and \textbf{CheXpert}. Out of these 14 disease classes, 5 of them exist in both datasets: Atelectasis, Cardiomegaly, Consolidation, Edema and Pleural effusion. The results we present are the \textbf{macro average} of the performance from both datasets.
    \item Unseen classes consist of one disease class, COVID-19, which solely comes from the \textbf{COVIDx CXR-4} dataset.
\end{itemize}
The seen classes help us identify the models' sensitivity across varied prompt styles, while the unseen classes are used to test the models' zero-shot inference ability on diseases not directly learned during pre-training. An ideal model should be both robust to prompt style variations on known diseases and able to effectively use highly interpretable prompts to improve predictions on unseen diseases.

In this study, we use the Area Under the Curve (AUC) as the main metric for evaluating model performance due to its ability to provide a comprehensive measure of the models' ability to discriminate between classes across all threshold levels. In addition to AUC, we also present F1 scores and accuracy (ACC) metrics in tables included in supplementary materials to provide a more rounded assessment of the models' performance across different prompt styles. In this section, we used abbreviations for prompt styles in graphs, namely: Disease Name -> Name, Disease Symptom -> Symptom, Disease Attribute -> Attribute, Disease Description(Plain English) -> Plain ENG, Disease Description(Medical Style) -> MED Style, Disease Description(Radiologist Style) -> RAD Style.

\textbf{For KAD, since the baseline style is identical to the Disease Name style, we only show the result for baseline style.}

\subsection{Performance on Seen Classes}

\begin{figure} [H]
  \centering
  \includegraphics[width=\linewidth]{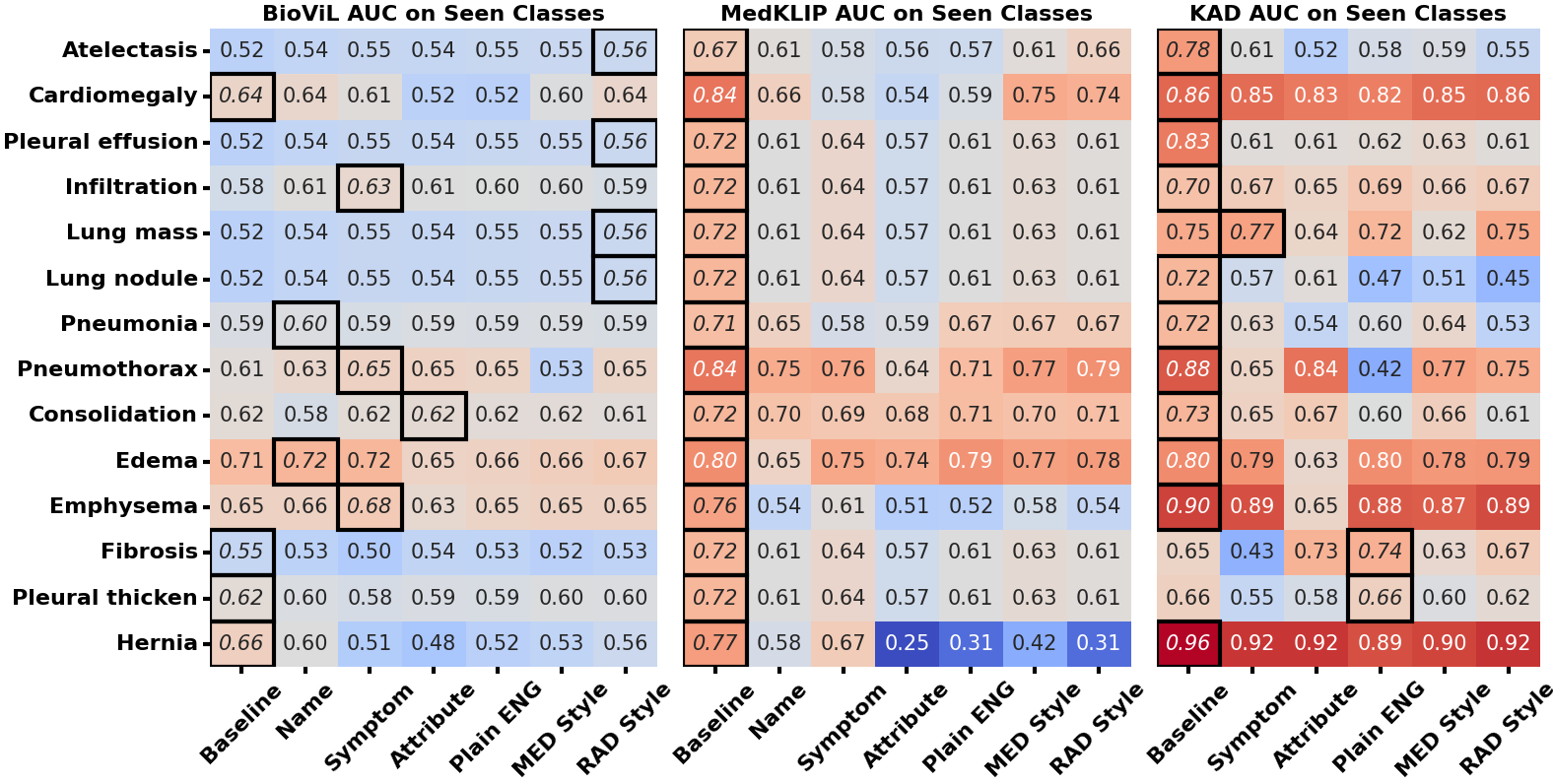}
  \caption{Heatmap demonstrating the performance of different models on \textbf{seen disease classes} with all prompt styles. The best performing prompt style of each disease class is highlighted with thick cell border and italic font.}
  \label{fig4}
\end{figure}

\begin{figure}
  \centering
  \includegraphics[width=\linewidth]{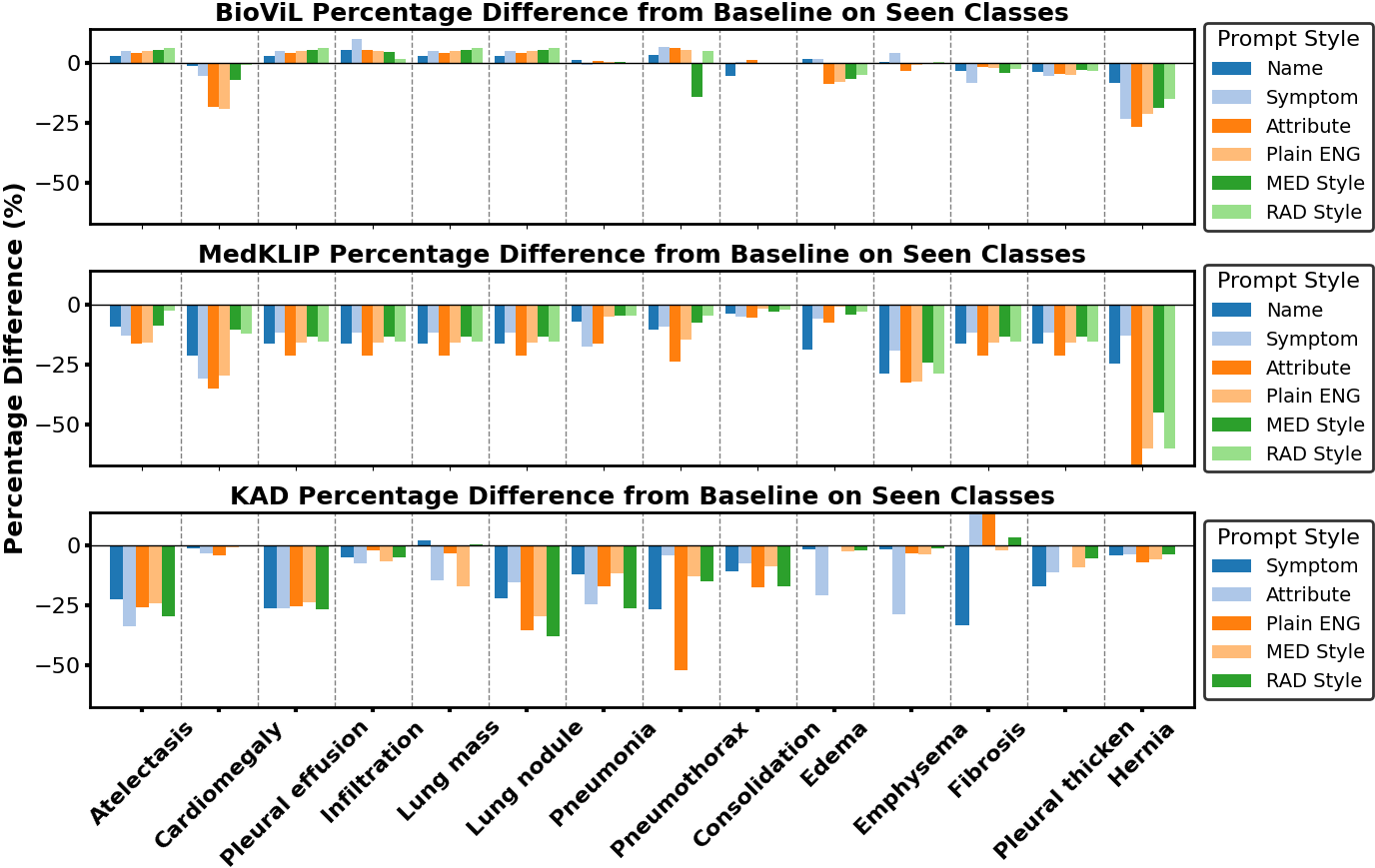}
  \caption{Bar charts demonstrating the performance difference of non-baseline prompt styles with baseline prompt style of different models on \textbf{seen disease classes}.}
  \label{fig5}
\end{figure}

In this section, we discuss the performance of the three mainstream MedVLP models on seen disease classes. To visualise the results, we provide three heatmaps in Figure\ref{fig4} displaying the AUC values, and three bar charts in Figure\ref{fig5} showing the percentage difference of non-baseline prompt styles with baseline prompt style.

BioViL demonstrates the most stable performance across various prompt styles compared to the other two models. However, it also exhibits an unremarkable overall performance, achieving a mean of 0.588 across all disease classes and prompt styles. The percentage differences in AUC between the baseline style and average of other prompt styles for BioViL range from a marginal improvement of +0.16\% for the Disease Description (Radiologist Style) to a slight decrease of -2.5\% for the Disease Attribute style.

MedKLIP, despite being designed to reduce dependency on specific prompt styles during training, shows a massive decrease in performance in all prompt styles that differed from the baseline style. The AUC for the Disease Description (Plain English) style, which is closest to the baseline style used in MedKLIP’s training, still shows a decrease of -18.48\%. The performance drops in non-baseline styles can be attributed to the model’s final pre-training step, where the baseline style prompts are directly encoded and used in contrastive learning between entity descriptions and CXR images.

KAD’s performance, while generally the most outstanding across the models, shows significant performance drops when tested with prompt styles different from the baseline style. The baseline style for KAD yields excellent results, reaching an average of 0.780 over all disease classes. The other prompt styles result in AUC decreases ranging from -11.2\% on Disease Description (Medical Style) to -13.38\% on Disease Attribute, highlighting the model's sensitivity to prompt style variation.

\subsection{Performance on Unseen Classes}

\begin{figure}[H]
  \centering
  \includegraphics[width=0.9\linewidth]{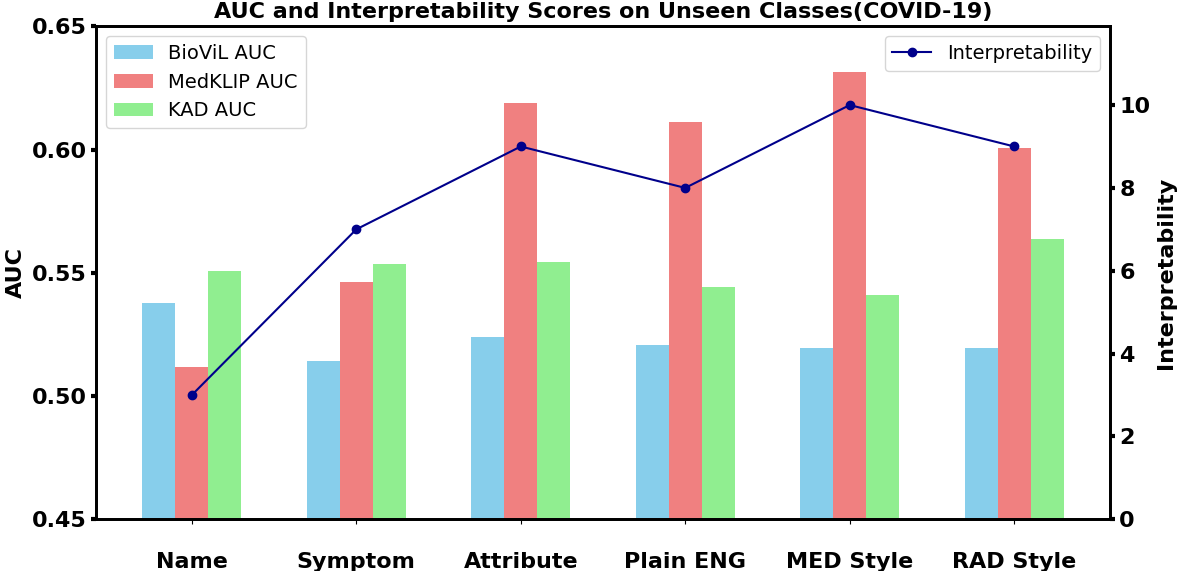}
  \caption{Graph showing both the interpretability rating of different prompt styles and the performance of different models on \textbf{unseen disease class(COVID-19)}. The right Y-axis shows the scale of interpretability scores of each prompt style. The left Y-axis shows the scale of the AUC scores.}
  \label{fig6}
\end{figure}

In this section, we discuss the performance of the three mainstream MedVLP models on unseen disease classes. We provide a bar/line chart in Figure\ref{fig6} to visualise the models' performance trend with varying prompt interpretability score.

For unseen classes, BioViL does not demonstrate the ability to leverage more detailed, informative prompts to learn new knowledge about these diseases. The performance is similar across different prompt styles, with no clear trend indicating improvement as prompt interpretability increases. For instance, the highest AUC achieved on COVID-19 class is 0.538 with Disease Name style while the other styles achieve only 0.524 or lower.

MedKLIP shows a significant ability to learn from information within prompts during inference on unseen classes. For COVID-19, the model's AUC increases from 0.511 (Disease Name style) to 0.631 (Disease Description Medical Style). It is also notable that the model's performance scales with the prompt styles' interpretability score, reflecting the model’s ability to utilise highly interpretable prompts effectively.

KAD shows some initial understanding of the disease class COVID-19, with AUC score using the simple disease names reaching 0.550. This observation can likely be explained with the fact that KAD's knowledge encoder was pre-trained with knowledge from UMLS database, which includes few pieces of introductions of this disease. However, KAD fails to demonstrate the ability to utilise high interpretability prompts, with all prompt styles yielding AUCs close to the baseline style (0.550 for Disease Name/baseline style and up to 0.564 for the Disease Description Radiologist style).

\subsection{Overall Analysis}
The results clearly highlight the main limitations of the current mainstream MedVLP models in handling variations in prompt styles during CXR zero-shot diagnosis.
\begin{itemize}[topsep=-3pt]
    \setlength{\itemsep}{1pt} 
    \setlength{\parskip}{1pt} 
    \setlength{\parsep}{1pt}  
    \item \textbf{BioViL:} While less sensitive to diverse prompt styles, BioViL delivers mediocre overall performance. It also shows minimal improvement when using high-interpretability prompts on unseen classes.
    \item \textbf{MedKLIP:} MedKLIP’s performance on seen classes is heavily influenced by the prompt styles used during pre-training, with severe performance drops observed when using non-baseline styles. However, MedKLIP does exhibit noticeable performance improvement on unseen diseases when provided with highly interpretable prompts.
    \item \textbf{KAD:} Despite its strong overall performance, KAD is highly sensitive to variations in prompt styles. It struggles with learning from more detailed prompts on unseen diseases and shows only marginal performance gains with increasing interpretability ratings.
\end{itemize} 

\subsection{Proposed Vision-Language Pre-training Recipe}
Based on our findings in Sections 5.1-5.3, we propose the following ideal recipe for future MedVLP model development:

\begin{itemize}[topsep=-3pt]
    \setlength{\itemsep}{1pt} 
    \setlength{\parskip}{1pt} 
    \setlength{\parsep}{1pt}  
    \item \textbf{Incorporate Domain Knowledge-Enhanced Approaches:} Both KAD \cite{KAD} and MedKLIP \cite{MedKLIP} utilise external knowledge databases, incorporating informative medical domain knowledge, which significantly enhances their zero-shot diagnosis performance. This is evidenced by their superior performance compared to BioViL, in both seen and unseen classes.
    \item \textbf{Pre-train with Informative Text:} MedKLIP’s scaling performance with prompt interpretability in unseen classes suggests that pre-training should incorporate prompts containing more descriptive and interpretable information. Including such informative text in the pre-training phase enables the model to better utilise the information in prompts during inference, thereby improving its performance.
    \item \textbf{Ensure Diverse Style of Text in Pre-training Dataset:} The performance inconsistency observed in MedKLIP and KAD when using baseline style prompts versus other styles underscores the importance of incorporating diverse text styles in the pre-training dataset. To enhance the model's adaptability and robustness during inference, it is crucial to include a broader range of prompt styles, ranging from simple disease names to detailed descriptions.
\end{itemize}

\section{Conclusion}
\label{sec:conclusion}
In this study, we conducted the first systematic evaluation of the sensitivity of three mainstream MedVLP methods to varying textual prompts across 15 distinct diseases in zero-shot classification tasks. We developed 6 unique prompt styles to replicate real-world clinical scenarios, ranked them by interpretability, and used these to assess the models' performance. Our analysis reveals that existing MedVLP models exhibit significant fluctuations in performance depending on the prompt styles, exposing a considerable gap in their robustness. Furthermore, the inconsistencies observed with more interpretable prompts suggest challenges in the models' ability to understand complex medical concepts. These findings underscore the need for further advancements in MedVLP techniques to improve their handling of diverse zero-shot prompts. We hope this work will inspire further research and innovation in the field of medical vision-language pre-training.

\newpage

\bibliographystyle{unsrt} 
\bibliography{references} 

\newpage
\appendix

\section{Appendix / supplemental material}
\setcounter{table}{0} 
\renewcommand{\tablename}{Supplementary Table} 

\begin{table}[htbp]
\centering
\renewcommand{\arraystretch}{1.25}
\caption{Construction prompts of diverse prompts Part A}
\begin{tabular}{|p{2cm}|p{10.5cm}|}
\hline
\textbf{Prompt Style} & \textbf{Construction Prompt} \\ \hline
\raggedright\textbf{Disease \newline Symptom} & You're a helpful AI radiologist. Help me enrich the simple disease names by adding their relevant symptoms that will help the diagnosis when only looking at the patient's chest x-ray image. \\
 & In other words, add the symptoms of the input disease that are commonly shown on the chest X-ray images of the patient if the patient is confirmed to have the input disease. \\
 & There are no limits on how many symptoms to add, but only include the ones that are most common. \\
 & Use [SEP] as the separator. \\
 & Let's think step by step. Start by listing all the possible symptoms in their most common names that can be shown on the chest x-ray image of the patient if the patient is confirmed with the input disease. Then pick the highly possible/common symptoms from them. Last, compile the highly possible symptoms, and put them into {}. \\
 & The text should be concise and follow the format in the examples below. \\
 & Here is one example. In your reply, only include content after "your output:". \\
 & """ \\
 & My input: Lung cancer \\
 & Your output: Possible symptoms: Nodule, Mass, Atelectasis, Pleural Effusion, Lymphadenopathy, Cavitation, Infiltrates, Rib Erosion. \\
 & Most common symptoms: Nodule, Mass, Atelectasis. \\
 & Final Output:{Lung cancer [SEP] Nodule [SEP] Mass [SEP] Atelectasis} \\
 & """ \\
 & Now process the following inputs: "atelectasis", "cardiomegaly", "pleural effusion", "infiltration", "lung mass", "lung nodule", "pneumonia", "pneumothorax", "consolidation", "edema", "emphysema", "fibrosis", "pleural thicken", "hernia", "COVID-19"\\ \hline
\end{tabular}
\label{tab10}
\end{table}

\begin{table}[htbp]
\centering
\renewcommand{\arraystretch}{1.25}
\caption{Construction prompts of diverse prompts Part B}
\begin{tabular}{|p{2cm}|p{10.5cm}|}
\hline
\textbf{Prompt Style} & \textbf{Construction Prompt} \\ \hline
\raggedright\textbf{Disease \newline Attribute} & You're a helpful AI radiologist. Help me enrich the simple disease names by adding their relevant attributes that will help the diagnosis when only looking at the patient's chest x-ray image. \\
 & Describe the disease from 8 visual attributes that describes the patient's chest X-ray image. \\
 & The 8 visual attributes are: border, fluid, location, opacity, other, patterns, shape, texture. \\
 & The text should be concise and follow the format in the examples below. \\
 & Here are two examples. In your reply, only include content after "Your output:". \\
 & """ \\
 & My input: normal \\
 & Your output: "border: clear and smooth, with the edge of the lung tissue appearing as a thin, curved line against the ribs.", \\
 & "fluid: no fluid or effusion accumulation.", \\
 & "location: fills the chest cavity, from just below the collarbones to just above the diaphragm.", \\
 & "opacity: balanced, neither too opaque (white) nor too transparent (dark).", \\
 & "other: symmetric appearance between two chest's sides; clear visibility of the heart, ribs, spine, and diaphragm; bronchial tubes and blood vessels are visible as white lines or tree-branch patterns against the darker lung tissue.", \\
 & "patterns: no cloudy or patchy areas, no concentrated white or black spots.", \\
 & "shape: lungs appear as two large, oval or triangular areas on either side of the heart.", \\
 & "texture: uniform with small, branching white lines representing the bronchi and blood vessels." \\
 & My input: effusion \\
 & Your output: "border: clear, sharp border along the top of the effusion.", \\
 & "fluid: fluid accumulation is the main feature, which causes a cloudy appearance.", \\
 & "location: typically located at the base of the lungs, between the lung and chest wall.", \\
 & "opacity: more opaque, appearing whiter or cloudier than the surrounding lung tissue.", \\
 & "other: possible displacement of other structures such as the heart or trachea; reduction in lung volume; and increased density at the base of the lung.", \\
 & "patterns: no specific patterns, effusion spreads out in the pleural space.", \\
 & "shape: typically appears as a meniscus, or curved shape, at the lung base.", \\
 & "texture: smooth texture, without any grainy or mottled appearance." \\
 & """ \\
 & Now process the following inputs: "atelectasis", "cardiomegaly", "pleural effusion", "infiltration", "lung mass", "lung nodule", "pneumonia", "pneumothorax", "consolidation", "edema", "emphysema", "fibrosis", "pleural thicken", "hernia", "COVID-19"\\ \hline
\end{tabular}
\label{tab11}
\end{table}

\begin{table}[htbp]
\centering
\renewcommand{\arraystretch}{1.25}
\caption{Construction prompts of diverse prompts Part C}
\begin{tabular}{|p{2cm}|p{10.5cm}|}
\hline
\textbf{Prompt Style} & \textbf{Construction Prompt} \\ \hline
\raggedright\textbf{Disease \newline Description (Plain \newline English)} & You're a helpful AI radiologist. Help me enrich the simple disease names by adding a concise but comprehensive description of the patient's chest X-ray image in plain English style. \\
 & Focus on what will be shown on the patient's Chest X-ray image. \\
 & The text should be concise and follow the format in the examples below. \\
 & Here are two examples. In your reply, only include content after "Your output:". \\
 & """ \\
 & My input: lung cancer \\
 & Your output: "The chest X-ray shows a suspicious mass in the left lung. This mass appears as a solid white area, indicating a potential tumor. The mass is approximately 4 centimeters in size and is located in the upper part of the left lung. There are also signs that the cancer may have spread to nearby lymph nodes, but no signs that it has spread to other parts of the body." \\
 & My input: effusion \\
 & Your output: "The chest X-ray shows an abnormal buildup of fluid in the space between the lungs and the chest wall, known as pleural effusion. This appears as a white area at the bottom of the lung. The fluid can cause the lung to partially collapse, making it harder to breathe." \\
 & Now process the following inputs: "atelectasis", "cardiomegaly", "pleural effusion", "infiltration", "lung mass", "lung nodule", "pneumonia", "pneumothorax", "consolidation", "edema", "emphysema", "fibrosis", "pleural thicken", "hernia", "COVID-19"\\ \hline
\raggedright\textbf{Disease \newline Description (Medical Style)} & You're a helpful AI radiologist. Help me enrich the simple disease names by adding a concise but comprehensive description of the patient's chest X-ray image in plain English style. \\
 & Focus on what will be shown on the patient's Chest X-ray image. \\
 & The text should be concise and follow the format in the examples below. \\
 & Here are two examples. In your reply, only include content after "Your output:". \\
 & """ \\
 & My input: lung cancer \\
 & Your output: "The chest X-ray shows a suspicious mass in the left lung. This mass appears as a solid white area, indicating a potential tumor. The mass is approximately 4 centimeters in size and is located in the upper part of the left lung. There are also signs that the cancer may have spread to nearby lymph nodes, but no signs that it has spread to other parts of the body." \\
 & My input: effusion \\
 & Your output: "The chest X-ray shows an abnormal buildup of fluid in the space between the lungs and the chest wall, known as pleural effusion. This appears as a white area at the bottom of the lung. The fluid can cause the lung to partially collapse, making it harder to breathe." \\
 & Now process the following inputs: "atelectasis", "cardiomegaly", "pleural effusion", "infiltration", "lung mass", "lung nodule", "pneumonia", "pneumothorax", "consolidation", "edema", "emphysema", "fibrosis", "pleural thicken", "hernia", "COVID-19"\\ \hline
\end{tabular}
\label{tab12}
\end{table}

\begin{table}[htbp]
\centering
\renewcommand{\arraystretch}{1.25}
\caption{Construction prompts of diverse prompts Part D}
\begin{tabular}{|p{2cm}|p{10.5cm}|}
\hline
\textbf{Prompt Style} & \textbf{Construction Prompt} \\ \hline
\raggedright\textbf{Disease \newline Description (Radiologist Style)} & You're a helpful AI radiologist. Help me enrich the simple disease names by adding a concise but comprehensive description of the patient's chest X-ray image in professional radiologist diagnosis style. \\
 & Focus on what will be shown on the patient's Chest X-ray image. \\
 & The text should be concise and follow the format in the examples below. \\
 & Here are two examples. In your reply, only include content after "Your output:". \\
 & """ \\
 & My input: lung cancer \\
 & Your output: "Radiographic findings on the chest X-ray demonstrate a 4 cm irregular, well-defined mass in the left upper lobe, presenting as an opaque region with spiculated margins. The lesion shows increased density compared to surrounding tissues. There is evidence of ipsilateral mediastinal lymphadenopathy. No pleural effusion or distant metastatic lesions are identified. Findings are consistent with stage IIIA non-small cell lung carcinoma (NSCLC). Further evaluation with contrast-enhanced CT and PET-CT is recommended for comprehensive staging and assessment." \\
 & My input: effusion \\
 & Your output: "Radiographic findings on the chest X-ray demonstrate a large pleural effusion in the right hemithorax, characterized by a homogenous opacity with a meniscus sign. There is a noticeable blunting of the costophrenic angle and partial atelectasis of the adjacent lung parenchyma. No evidence of pneumothorax or significant mediastinal shift is observed. Further evaluation with ultrasound or contrast-enhanced CT is recommended to assess the nature and extent of the effusion." \\
 & Now process the following inputs: "atelectasis", "cardiomegaly", "pleural effusion", "infiltration", "lung mass", "lung nodule", "pneumonia", "pneumothorax", "consolidation", "edema", "emphysema", "fibrosis", "pleural thicken", "hernia", "COVID-19"\\ \hline
\end{tabular}
\label{tab13}
\end{table}

\begin{table}[htbp]
\centering
\renewcommand{\arraystretch}{1.25}
\caption{Interpretability rating prompt}
\begin{tabular}{|p{13cm}|}
\hline
\textbf{Interpretability Rating Prompt} \\ \hline
Below are 6 different styles to describe a disease. \\
Rate the 6 styles in terms of interpretability with scale of 1 to 10. \\
10 being the most interpretable and the most informative for diagnosing the disease, and 1 being the least interpretable and the least informative. \\
\textit{Following are the diverse prompts for all disease classes and prompt styles. Due to length issue, we omit them from this table. For all diverse prompts, see the supplementary materials of this paper.} \\ \hline
\end{tabular}
\label{tab14}
\end{table}

\clearpage

\begin{table}[htbp]
\centering
\renewcommand{\arraystretch}{1.25}
\caption{Results of zero-shot image classification on the \textbf{ChestX-ray14} dataset with the \textbf{BioViL} model. The best performing prompt style for each disease class is highlighted in bold.}
\begin{tabular}{|c|c|ccccccc|}
\hline
\rotatebox{90}{\textbf{Metric}} & \textbf{Disease Class} & \rotatebox{90}{\textbf{Baseline}} & \rotatebox{90}{\textbf{Name}} & \rotatebox{90}{\textbf{Symptom}} & \rotatebox{90}{\textbf{Attribute}} & \rotatebox{90}{\textbf{Plain ENG}} & \rotatebox{90}{\textbf{MED Style}} & \rotatebox{90}{\textbf{RAD Style}}\\
\hline
\multirow{14}{*}{AUC} & atelectasis & 0.524 & 0.539 & 0.549 & 0.545 & 0.550 & 0.553 & \textbf{0.555} \\
 & cardiomegaly & \textbf{0.645} & 0.637 & 0.610 & 0.525 & 0.521 & 0.600 & 0.639 \\
 & pleural effusion & 0.524 & 0.539 & 0.549 & 0.545 & 0.550 & 0.553 & \textbf{0.555} \\
 & infiltration & 0.576 & 0.607 & \textbf{0.634} & 0.606 & 0.603 & 0.601 & 0.586 \\
 & lung mass & 0.524 & 0.539 & 0.549 & 0.545 & 0.550 & 0.553 & \textbf{0.555} \\
 & lung nodule & 0.524 & 0.539 & 0.549 & 0.545 & 0.550 & 0.553 & \textbf{0.555} \\
 & pneumonia & 0.591 & \textbf{0.597} & 0.586 & 0.594 & 0.592 & 0.593 & 0.590 \\
 & pneumothorax & 0.615 & 0.635 & \textbf{0.654} & 0.653 & 0.646 & 0.528 & 0.645 \\
 & consolidation & 0.618 & 0.584 & 0.619 & \textbf{0.624} & 0.616 & 0.616 & 0.614 \\
 & edema & 0.713 & \textbf{0.725} & 0.724 & 0.651 & 0.656 & 0.663 & 0.675 \\
 & emphysema & 0.653 & 0.656 & \textbf{0.678} & 0.629 & 0.647 & 0.653 & 0.654 \\
 & fibrosis & \textbf{0.546} & 0.527 & 0.500 & 0.536 & 0.534 & 0.521 & 0.532 \\
 & pleural thicken & \textbf{0.619} & 0.595 & 0.584 & 0.591 & 0.589 & 0.599 & 0.597 \\
 & hernia & \textbf{0.659} & 0.605 & 0.506 & 0.484 & 0.519 & 0.535 & 0.559 \\
\hline
\multirow{14}{*}{F1} & atelectasis & 0.224 & 0.231 & 0.242 & 0.240 & 0.243 & 0.244 & \textbf{0.246} \\
 & cardiomegaly & 0.125 & 0.122 & 0.112 & 0.084 & 0.083 & 0.106 & \textbf{0.129} \\
 & pleural effusion & 0.224 & 0.231 & 0.242 & 0.240 & 0.243 & 0.244 & \textbf{0.246} \\
 & infiltration & 0.406 & 0.432 & \textbf{0.458} & 0.438 & 0.436 & 0.432 & 0.421 \\
 & lung mass & 0.224 & 0.231 & 0.242 & 0.240 & 0.243 & 0.244 & \textbf{0.246} \\
 & lung nodule & 0.224 & 0.231 & 0.242 & 0.240 & 0.243 & 0.244 & \textbf{0.246} \\
 & pneumonia & 0.052 & 0.053 & 0.052 & 0.053 & 0.053 & \textbf{0.053} & 0.052 \\
 & pneumothorax & 0.234 & 0.243 & 0.254 & \textbf{0.254} & 0.253 & 0.189 & 0.252 \\
 & consolidation & 0.168 & 0.157 & 0.169 & \textbf{0.171} & 0.168 & 0.169 & 0.167 \\
 & edema & 0.123 & 0.129 & \textbf{0.129} & 0.100 & 0.102 & 0.104 & 0.108 \\
 & emphysema & 0.119 & 0.118 & \textbf{0.126} & 0.109 & 0.114 & 0.116 & 0.116 \\
 & fibrosis & \textbf{0.038} & 0.036 & 0.033 & 0.037 & 0.036 & 0.035 & 0.036 \\
 & pleural thicken & \textbf{0.116} & 0.106 & 0.103 & 0.105 & 0.104 & 0.108 & 0.107 \\
 & hernia & \textbf{0.012} & 0.009 & 0.007 & 0.006 & 0.007 & 0.007 & 0.008 \\
\hline
\multirow{14}{*}{ACC} & atelectasis & 0.463 & \textbf{0.536} & 0.409 & 0.398 & 0.398 & 0.435 & 0.405 \\
 & cardiomegaly & 0.581 & 0.579 & 0.561 & 0.339 & 0.334 & 0.495 & \textbf{0.640} \\
 & pleural effusion & 0.463 & \textbf{0.536} & 0.409 & 0.398 & 0.398 & 0.435 & 0.405 \\
 & infiltration & 0.511 & \textbf{0.546} & 0.536 & 0.476 & 0.477 & 0.495 & 0.473 \\
 & lung mass & 0.463 & \textbf{0.536} & 0.409 & 0.398 & 0.398 & 0.435 & 0.405 \\
 & lung nodule & 0.463 & \textbf{0.536} & 0.409 & 0.398 & 0.398 & 0.435 & 0.405 \\
 & pneumonia & 0.283 & 0.292 & 0.276 & 0.299 & 0.286 & \textbf{0.323} & 0.282 \\
 & pneumothorax & 0.382 & 0.383 & 0.414 & 0.423 & 0.446 & \textbf{0.512} & 0.448 \\
 & consolidation & 0.322 & 0.323 & 0.344 & \textbf{0.364} & 0.338 & 0.360 & 0.323 \\
 & edema & 0.530 & 0.556 & \textbf{0.561} & 0.412 & 0.420 & 0.439 & 0.451 \\
 & emphysema & 0.443 & 0.418 & \textbf{0.454} & 0.343 & 0.366 & 0.378 & 0.379 \\
 & fibrosis & 0.332 & 0.359 & \textbf{0.367} & 0.300 & 0.300 & 0.350 & 0.340 \\
 & pleural thicken & \textbf{0.455} & 0.346 & 0.314 & 0.333 & 0.329 & 0.366 & 0.352 \\
 & hernia & \textbf{0.587} & 0.489 & 0.338 & 0.283 & 0.308 & 0.338 & 0.363 \\
\hline
\end{tabular}
\label{tab1}
\end{table}

\begin{table}[htbp]
\centering
\renewcommand{\arraystretch}{1.25}
\caption{Results of zero-shot image classification on the \textbf{ChestX-ray14} dataset with the \textbf{MedKLIP} model. The best performing prompt style for each disease class is highlighted in bold.}
\begin{tabular}{|c|c|ccccccc|}
\hline
\rotatebox{90}{\textbf{Metric}} & \textbf{Disease Class} & \rotatebox{90}{\textbf{Baseline}} & \rotatebox{90}{\textbf{Name}} & \rotatebox{90}{\textbf{Symptom}} & \rotatebox{90}{\textbf{Attribute}} & \rotatebox{90}{\textbf{Plain ENG}} & \rotatebox{90}{\textbf{MED Style}} & \rotatebox{90}{\textbf{RAD Style}}\\
\hline
\multirow{14}{*}{AUC} & atelectasis & \textbf{0.673} & 0.611 & 0.584 & 0.562 & 0.565 & 0.613 & 0.655 \\
 & cardiomegaly & \textbf{0.839} & 0.660 & 0.579 & 0.545 & 0.589 & 0.749 & 0.737 \\
 & pleural effusion & \textbf{0.723} & 0.606 & 0.639 & 0.569 & 0.606 & 0.625 & 0.611 \\
 & infiltration & \textbf{0.723} & 0.606 & 0.639 & 0.569 & 0.606 & 0.625 & 0.611 \\
 & lung mass & \textbf{0.723} & 0.606 & 0.639 & 0.569 & 0.606 & 0.625 & 0.611 \\
 & lung nodule & \textbf{0.723} & 0.606 & 0.639 & 0.569 & 0.606 & 0.625 & 0.611 \\
 & pneumonia & \textbf{0.707} & 0.654 & 0.582 & 0.592 & 0.669 & 0.672 & 0.672 \\
 & pneumothorax & \textbf{0.836} & 0.745 & 0.757 & 0.636 & 0.713 & 0.770 & 0.795 \\
 & consolidation & \textbf{0.723} & 0.696 & 0.686 & 0.682 & 0.710 & 0.700 & 0.705 \\
 & edema & \textbf{0.802} & 0.649 & 0.754 & 0.741 & 0.793 & 0.768 & 0.776 \\
 & emphysema & \textbf{0.761} & 0.541 & 0.615 & 0.513 & 0.515 & 0.575 & 0.542 \\
 & fibrosis & \textbf{0.723} & 0.606 & 0.639 & 0.569 & 0.606 & 0.625 & 0.611 \\
 & pleural thicken & \textbf{0.723} & 0.606 & 0.639 & 0.569 & 0.606 & 0.625 & 0.611 \\
 & hernia & \textbf{0.772} & 0.581 & 0.672 & 0.253 & 0.307 & 0.423 & 0.306 \\
\hline
\multirow{14}{*}{F1} & atelectasis & \textbf{0.299} & 0.265 & 0.255 & 0.259 & 0.265 & 0.272 & 0.286 \\
 & cardiomegaly & \textbf{0.295} & 0.128 & 0.097 & 0.089 & 0.096 & 0.205 & 0.185 \\
 & pleural effusion & \textbf{0.254} & 0.173 & 0.187 & 0.168 & 0.180 & 0.194 & 0.191 \\
 & infiltration & \textbf{0.254} & 0.173 & 0.187 & 0.168 & 0.180 & 0.194 & 0.191 \\
 & lung mass & \textbf{0.254} & 0.173 & 0.187 & 0.168 & 0.180 & 0.194 & 0.191 \\
 & lung nodule & \textbf{0.254} & 0.173 & 0.187 & 0.168 & 0.180 & 0.194 & 0.191 \\
 & pneumonia & \textbf{0.100} & 0.081 & 0.054 & 0.056 & 0.076 & 0.078 & 0.078 \\
 & pneumothorax & \textbf{0.453} & 0.339 & 0.359 & 0.237 & 0.284 & 0.378 & 0.400 \\
 & consolidation & \textbf{0.225} & 0.207 & 0.204 & 0.199 & 0.208 & 0.210 & 0.211 \\
 & edema & \textbf{0.193} & 0.107 & 0.146 & 0.144 & 0.178 & 0.172 & 0.175 \\
 & emphysema & \textbf{0.296} & 0.095 & 0.112 & 0.082 & 0.091 & 0.099 & 0.094 \\
 & fibrosis & \textbf{0.254} & 0.173 & 0.187 & 0.168 & 0.180 & 0.194 & 0.191 \\
 & pleural thicken & \textbf{0.254} & 0.173 & 0.187 & 0.168 & 0.180 & 0.194 & 0.191 \\
 & hernia & \textbf{0.100} & 0.011 & 0.016 & 0.007 & 0.007 & 0.007 & 0.007 \\
\hline
\multirow{14}{*}{ACC} & atelectasis & 0.596 & 0.508 & 0.427 & 0.393 & 0.417 & 0.449 & \textbf{0.621} \\
 & cardiomegaly & 0.914 & 0.804 & 0.446 & 0.322 & 0.356 & \textbf{0.921} & 0.897 \\
 & pleural effusion & \textbf{0.808} & 0.581 & 0.603 & 0.441 & 0.508 & 0.536 & 0.553 \\
 & infiltration & \textbf{0.808} & 0.581 & 0.603 & 0.441 & 0.508 & 0.536 & 0.553 \\
 & lung mass & \textbf{0.808} & 0.581 & 0.603 & 0.441 & 0.508 & 0.536 & 0.553 \\
 & lung nodule & \textbf{0.808} & 0.581 & 0.603 & 0.441 & 0.508 & 0.536 & 0.553 \\
 & pneumonia & 0.883 & \textbf{0.925} & 0.509 & 0.646 & 0.836 & 0.802 & 0.852 \\
 & pneumothorax & \textbf{0.851} & 0.800 & 0.816 & 0.475 & 0.776 & 0.851 & 0.850 \\
 & consolidation & 0.659 & 0.674 & \textbf{0.743} & 0.611 & 0.639 & 0.619 & 0.705 \\
 & edema & 0.878 & 0.790 & 0.742 & 0.811 & 0.860 & \textbf{0.880} & 0.858 \\
 & emphysema & \textbf{0.941} & 0.332 & 0.646 & 0.086 & 0.253 & 0.389 & 0.311 \\
 & fibrosis & \textbf{0.808} & 0.581 & 0.603 & 0.441 & 0.508 & 0.536 & 0.553 \\
 & pleural thicken & \textbf{0.808} & 0.581 & 0.603 & 0.441 & 0.508 & 0.536 & 0.553 \\
 & hernia & \textbf{0.991} & 0.809 & 0.842 & 0.047 & 0.068 & 0.120 & 0.009 \\
\hline
\end{tabular}
\label{tab2}
\end{table}

\begin{table}[htbp]
\centering
\renewcommand{\arraystretch}{1.25}
\caption{Results of zero-shot image classification on the \textbf{ChestX-ray14} dataset with the \textbf{KAD} model. The best performing prompt style for each disease class is highlighted in bold.}
\begin{tabular}{|c|c|cccccc|}
\hline
\rotatebox{90}{\textbf{Metric}} & \textbf{Disease Class} & \rotatebox{90}{\textbf{Baseline}} & \rotatebox{90}{\textbf{Symptom}} & \rotatebox{90}{\textbf{Attribute}} & \rotatebox{90}{\textbf{Plain ENG}} & \rotatebox{90}{\textbf{MED Style}} & \rotatebox{90}{\textbf{RAD Style}}\\
\hline
\multirow{14}{*}{AUC} & atelectasis & \textbf{0.779} & 0.606 & 0.516 & 0.581 & 0.592 & 0.549 \\
 & cardiomegaly & \textbf{0.858} & 0.850 & 0.831 & 0.822 & 0.851 & 0.855 \\
 & pleural effusion & \textbf{0.828} & 0.611 & 0.613 & 0.620 & 0.632 & 0.610 \\
 & infiltration & \textbf{0.700} & 0.666 & 0.648 & 0.687 & 0.655 & 0.665 \\
 & lung mass & 0.749 & \textbf{0.765} & 0.642 & 0.724 & 0.623 & 0.754 \\
 & lung nodule & \textbf{0.725} & 0.567 & 0.613 & 0.470 & 0.510 & 0.450 \\
 & pneumonia & \textbf{0.717} & 0.631 & 0.543 & 0.596 & 0.636 & 0.531 \\
 & pneumothorax & \textbf{0.877} & 0.646 & 0.841 & 0.421 & 0.766 & 0.746 \\
 & consolidation & \textbf{0.727} & 0.651 & 0.675 & 0.602 & 0.664 & 0.606 \\
 & edema & \textbf{0.802} & 0.789 & 0.635 & 0.801 & 0.782 & 0.788 \\
 & emphysema & \textbf{0.904} & 0.889 & 0.645 & 0.875 & 0.871 & 0.893 \\
 & fibrosis & 0.647 & 0.433 & 0.731 & \textbf{0.737} & 0.633 & 0.671 \\
 & pleural thicken & 0.657 & 0.546 & 0.583 & \textbf{0.659} & 0.597 & 0.624 \\
 & hernia & \textbf{0.955} & 0.918 & 0.919 & 0.888 & 0.903 & 0.922 \\
\hline
\multirow{14}{*}{F1} & atelectasis & \textbf{0.402} & 0.266 & 0.237 & 0.258 & 0.259 & 0.241 \\
 & cardiomegaly & 0.362 & \textbf{0.366} & 0.346 & 0.305 & 0.366 & 0.362 \\
 & pleural effusion & \textbf{0.537} & 0.350 & 0.349 & 0.359 & 0.365 & 0.346 \\
 & infiltration & \textbf{0.481} & 0.458 & 0.444 & 0.467 & 0.447 & 0.453 \\
 & lung mass & 0.297 & \textbf{0.314} & 0.183 & 0.281 & 0.165 & 0.286 \\
 & lung nodule & \textbf{0.262} & 0.134 & 0.153 & 0.122 & 0.121 & 0.122 \\
 & pneumonia & \textbf{0.082} & 0.057 & 0.048 & 0.049 & 0.058 & 0.047 \\
 & pneumothorax & \textbf{0.505} & 0.245 & 0.434 & 0.201 & 0.350 & 0.321 \\
 & consolidation & \textbf{0.215} & 0.177 & 0.195 & 0.151 & 0.187 & 0.159 \\
 & edema & 0.160 & 0.154 & 0.092 & \textbf{0.165} & 0.150 & 0.163 \\
 & emphysema & \textbf{0.479} & 0.445 & 0.117 & 0.384 & 0.382 & 0.458 \\
 & fibrosis & 0.064 & 0.035 & 0.070 & \textbf{0.084} & 0.050 & 0.058 \\
 & pleural thicken & \textbf{0.136} & 0.102 & 0.094 & 0.134 & 0.105 & 0.128 \\
 & hernia & \textbf{0.537} & 0.504 & 0.510 & 0.403 & 0.531 & 0.530 \\
\hline
\multirow{14}{*}{ACC} & atelectasis & \textbf{0.815} & 0.393 & 0.236 & 0.364 & 0.391 & 0.210 \\
 & cardiomegaly & 0.938 & 0.932 & \textbf{0.939} & 0.929 & 0.938 & 0.928 \\
 & pleural effusion & \textbf{0.812} & 0.435 & 0.385 & 0.420 & 0.443 & 0.369 \\
 & infiltration & \textbf{0.670} & 0.589 & 0.514 & 0.622 & 0.548 & 0.546 \\
 & lung mass & 0.904 & \textbf{0.905} & 0.766 & 0.893 & 0.568 & 0.896 \\
 & lung nodule & \textbf{0.909} & 0.437 & 0.588 & 0.106 & 0.148 & 0.134 \\
 & pneumonia & \textbf{0.681} & 0.467 & 0.317 & 0.224 & 0.452 & 0.148 \\
 & pneumothorax & 0.858 & 0.564 & 0.815 & 0.190 & \textbf{0.865} & 0.778 \\
 & consolidation & \textbf{0.603} & 0.474 & 0.566 & 0.249 & 0.547 & 0.314 \\
 & edema & 0.704 & 0.672 & 0.366 & 0.723 & 0.689 & \textbf{0.739} \\
 & emphysema & \textbf{0.957} & 0.955 & 0.460 & 0.946 & 0.947 & 0.952 \\
 & fibrosis & \textbf{0.819} & 0.063 & 0.700 & 0.815 & 0.612 & 0.667 \\
 & pleural thicken & 0.663 & \textbf{0.851} & 0.200 & 0.650 & 0.461 & 0.789 \\
 & hernia & 0.997 & \textbf{0.997} & 0.997 & 0.997 & 0.997 & 0.997 \\
\hline
\end{tabular}
\label{tab3}
\end{table}

\clearpage

\begin{table}[htbp]
\centering
\renewcommand{\arraystretch}{1.25}
\caption{Results of zero-shot image classification on the \textbf{CheXpert} dataset with the \textbf{BioViL} model. The best performing prompt style for each disease class is highlighted in bold.}
\begin{tabular}{|c|c|ccccccc|}
\hline
\rotatebox{90}{\textbf{Metric}} & \textbf{Disease Class} & \rotatebox{90}{\textbf{Baseline}} & \rotatebox{90}{\textbf{Name}} & \rotatebox{90}{\textbf{Symptom}} & \rotatebox{90}{\textbf{Attribute}} & \rotatebox{90}{\textbf{Plain ENG}} & \rotatebox{90}{\textbf{MED Style}} & \rotatebox{90}{\textbf{RAD Style}}\\
\hline
\multirow{5}{*}{AUC} & atelectasis & 0.646 & 0.644 & 0.750 & 0.753 & 0.751 & 0.750 & \textbf{0.757} \\
 & cardiomegaly & \textbf{0.720} & 0.711 & 0.687 & 0.684 & 0.718 & 0.694 & 0.690 \\
 & consolidation & 0.709 & 0.636 & 0.733 & \textbf{0.749} & 0.720 & 0.647 & 0.677 \\
 & edema & 0.687 & 0.698 & 0.701 & \textbf{0.703} & 0.667 & 0.682 & 0.681 \\
 & pleural effusion & 0.777 & 0.782 & 0.765 & 0.787 & \textbf{0.799} & 0.755 & 0.764 \\
\hline
\multirow{5}{*}{F1} & atelectasis & 0.496 & 0.495 & 0.611 & 0.615 & 0.612 & 0.610 & \textbf{0.619} \\
 & cardiomegaly & \textbf{0.573} & 0.561 & 0.538 & 0.532 & 0.568 & 0.545 & 0.537 \\
 & consolidation & 0.171 & 0.141 & 0.192 & \textbf{0.205} & 0.182 & 0.148 & 0.157 \\
 & edema & 0.442 & 0.457 & \textbf{0.464} & 0.427 & 0.382 & 0.396 & 0.412 \\
 & pleural effusion & 0.522 & 0.524 & 0.500 & 0.535 & \textbf{0.557} & 0.541 & 0.502 \\
\hline
\multirow{5}{*}{ACC} & atelectasis & 0.623 & 0.603 & 0.735 & \textbf{0.738} & 0.732 & 0.731 & 0.738 \\
 & cardiomegaly & 0.714 & 0.701 & \textbf{0.763} & 0.689 & 0.690 & 0.720 & 0.654 \\
 & consolidation & 0.551 & 0.490 & 0.623 & \textbf{0.653} & 0.597 & 0.536 & 0.516 \\
 & edema & 0.849 & 0.850 & \textbf{0.855} & 0.807 & 0.796 & 0.795 & 0.820 \\
 & pleural effusion & 0.710 & 0.707 & 0.674 & 0.725 & 0.750 & \textbf{0.784} & 0.683 \\
\hline
\end{tabular}
\label{tab4}
\end{table}

\begin{table}[htbp]
\centering
\renewcommand{\arraystretch}{1.25}
\caption{Results of zero-shot image classification on the \textbf{CheXpert} dataset with the \textbf{MedKLIP} model. The best performing prompt style for each disease class is highlighted in bold.}
\begin{tabular}{|c|c|ccccccc|}
\hline
\rotatebox{90}{\textbf{Metric}} & \textbf{Disease Class} & \rotatebox{90}{\textbf{Baseline}} & \rotatebox{90}{\textbf{Name}} & \rotatebox{90}{\textbf{Symptom}} & \rotatebox{90}{\textbf{Attribute}} & \rotatebox{90}{\textbf{Plain ENG}} & \rotatebox{90}{\textbf{MED Style}} & \rotatebox{90}{\textbf{RAD Style}}\\
\hline
\multirow{5}{*}{AUC} & atelectasis & \textbf{0.870} & 0.844 & 0.825 & 0.811 & 0.839 & 0.838 & 0.787 \\
 & cardiomegaly & \textbf{0.899} & 0.802 & 0.771 & 0.417 & 0.761 & 0.816 & 0.806 \\
 & consolidation & 0.897 & 0.896 & 0.816 & 0.852 & \textbf{0.905} & 0.805 & 0.856 \\
 & edema & \textbf{0.924} & 0.685 & 0.775 & 0.770 & 0.902 & 0.845 & 0.895 \\
 & pleural effusion & \textbf{0.909} & 0.801 & 0.821 & 0.739 & 0.865 & 0.848 & 0.847 \\
\hline
\multirow{5}{*}{F1} & atelectasis & \textbf{0.682} & 0.663 & 0.639 & 0.649 & 0.652 & 0.649 & 0.584 \\
 & cardiomegaly & \textbf{0.706} & 0.573 & 0.577 & 0.427 & 0.563 & 0.610 & 0.603 \\
 & consolidation & \textbf{0.456} & 0.434 & 0.323 & 0.425 & 0.411 & 0.333 & 0.360 \\
 & edema & \textbf{0.621} & 0.313 & 0.413 & 0.404 & 0.575 & 0.503 & 0.544 \\
 & pleural effusion & \textbf{0.648} & 0.493 & 0.529 & 0.499 & 0.574 & 0.565 & 0.549 \\
\hline
\multirow{5}{*}{ACC} & atelectasis & \textbf{0.811} & 0.790 & 0.789 & 0.796 & 0.768 & 0.765 & 0.678 \\
 & cardiomegaly & \textbf{0.819} & 0.744 & 0.722 & 0.337 & 0.713 & 0.759 & 0.714 \\
 & consolidation & \textbf{0.934} & 0.909 & 0.867 & 0.930 & 0.934 & 0.915 & 0.913 \\
 & edema & 0.882 & 0.723 & 0.807 & 0.831 & 0.855 & 0.868 & \textbf{0.898} \\
 & pleural effusion & \textbf{0.873} & 0.786 & 0.811 & 0.742 & 0.831 & 0.842 & 0.811 \\
\hline
\end{tabular}
\label{tab5}
\end{table}

\begin{table}[htbp]
\centering
\renewcommand{\arraystretch}{1.25}
\caption{Results of zero-shot image classification on the \textbf{CheXpert} dataset with the \textbf{KAD} model. The best performing prompt style for each disease class is highlighted in bold.}
\begin{tabular}{|c|c|cccccc|}
\hline
\rotatebox{90}{\textbf{Metric}} & \textbf{Disease Class} & \rotatebox{90}{\textbf{Baseline}} & \rotatebox{90}{\textbf{Symptom}} & \rotatebox{90}{\textbf{Attribute}} & \rotatebox{90}{\textbf{Plain ENG}} & \rotatebox{90}{\textbf{MED Style}} & \rotatebox{90}{\textbf{RAD Style}}\\
\hline
\multirow{5}{*}{AUC} & atelectasis & \textbf{0.847} & 0.728 & 0.626 & 0.765 & 0.750 & 0.730 \\
 & cardiomegaly & 0.860 & 0.867 & 0.853 & 0.776 & 0.844 & \textbf{0.871} \\
 & consolidation & \textbf{0.867} & 0.702 & 0.697 & 0.768 & 0.728 & 0.713 \\
 & edema & \textbf{0.932} & 0.904 & 0.714 & 0.911 & 0.877 & 0.884 \\
 & pleural effusion & \textbf{0.964} & 0.701 & 0.783 & 0.755 & 0.766 & 0.765 \\
\hline
\multirow{5}{*}{F1} & atelectasis & \textbf{0.637} & 0.524 & 0.466 & 0.557 & 0.541 & 0.550 \\
 & cardiomegaly & 0.658 & 0.677 & 0.655 & 0.568 & 0.632 & \textbf{0.680} \\
 & consolidation & 0.250 & 0.275 & 0.148 & 0.277 & 0.234 & \textbf{0.314} \\
 & edema & \textbf{0.667} & 0.593 & 0.298 & 0.601 & 0.516 & 0.544 \\
 & pleural effusion & \textbf{0.783} & 0.416 & 0.474 & 0.449 & 0.451 & 0.453 \\
\hline
\multirow{5}{*}{ACC} & atelectasis & \textbf{0.794} & 0.628 & 0.527 & 0.650 & 0.741 & 0.622 \\
 & cardiomegaly & 0.764 & \textbf{0.805} & 0.752 & 0.695 & 0.723 & 0.803 \\
 & consolidation & \textbf{0.953} & 0.942 & 0.425 & 0.845 & 0.867 & 0.945 \\
 & edema & \textbf{0.900} & 0.867 & 0.433 & 0.873 & 0.806 & 0.838 \\
 & pleural effusion & \textbf{0.922} & 0.600 & 0.730 & 0.770 & 0.647 & 0.766 \\
\hline
\end{tabular}
\label{tab6}
\end{table}

\clearpage

\begin{table}[htbp]
\centering
\renewcommand{\arraystretch}{1.25}
\caption{Results of zero-shot image classification on the \textbf{COVIDx CXR-4} dataset with the \textbf{BioViL} model. The best performing prompt style for each disease class is highlighted in bold.}
\begin{tabular}{|c|c|ccccccc|}
\hline
\rotatebox{90}{\textbf{Metric}} & \textbf{Disease Class} & \rotatebox{90}{\textbf{Baseline}} & \rotatebox{90}{\textbf{Name}} & \rotatebox{90}{\textbf{Symptom}} & \rotatebox{90}{\textbf{Attribute}} & \rotatebox{90}{\textbf{Plain ENG}} & \rotatebox{90}{\textbf{MED Style}} & \rotatebox{90}{\textbf{RAD Style}}\\
\hline
\multirow{1}{*}{AUC} & COVID-19 & 0.499 & \textbf{0.538} & 0.514 & 0.524 & 0.521 & 0.520 & 0.520 \\
\hline
\multirow{1}{*}{F1} & COVID-19 & 0.467 & 0.507 & 0.480 & 0.472 & 0.513 & 0.495 & \textbf{0.531} \\
\hline
\multirow{1}{*}{ACC} & COVID-19 & 0.499 & \textbf{0.538} & 0.514 & 0.524 & 0.521 & 0.520 & 0.520 \\
\hline
\end{tabular}
\label{tab7}
\end{table}

\begin{table}[htbp]
\centering
\renewcommand{\arraystretch}{1.25}
\caption{Results of zero-shot image classification on the \textbf{COVIDx CXR-4} dataset with the \textbf{MedKLIP} model. The best performing prompt style for each disease class is highlighted in bold.}
\begin{tabular}{|c|c|ccccccc|}
\hline
\rotatebox{90}{\textbf{Metric}} & \textbf{Disease Class} & \rotatebox{90}{\textbf{Baseline}} & \rotatebox{90}{\textbf{Name}} & \rotatebox{90}{\textbf{Symptom}} & \rotatebox{90}{\textbf{Attribute}} & \rotatebox{90}{\textbf{Plain ENG}} & \rotatebox{90}{\textbf{MED Style}} & \rotatebox{90}{\textbf{RAD Style}}\\
\hline
\multirow{1}{*}{AUC} & COVID-19 & 0.594 & 0.512 & 0.546 & 0.619 & 0.611 & \textbf{0.631} & 0.600 \\
\hline
\multirow{1}{*}{F1} & COVID-19 & \textbf{0.677} & 0.667 & 0.668 & 0.674 & 0.668 & 0.675 & 0.667 \\
\hline
\multirow{1}{*}{ACC} & COVID-19 & \textbf{0.561} & 0.501 & 0.507 & 0.548 & 0.516 & 0.536 & 0.502 \\
\hline
\end{tabular}
\label{table8}
\end{table}

\begin{table}[htbp]
\centering
\renewcommand{\arraystretch}{1.25}
\caption{Results of zero-shot image classification on the \textbf{COVIDx CXR-4} dataset with the \textbf{KAD} model. The best performing prompt style for each disease class is highlighted in bold.}
\begin{tabular}{|c|c|cccccc|}
\hline
\rotatebox{90}{\textbf{Metric}} & \textbf{Disease Class} & \rotatebox{90}{\textbf{Baseline}} & \rotatebox{90}{\textbf{Symptom}} & \rotatebox{90}{\textbf{Attribute}} & \rotatebox{90}{\textbf{Plain ENG}} & \rotatebox{90}{\textbf{MED Style}} & \rotatebox{90}{\textbf{RAD Style}}\\
\hline
\multirow{1}{*}{AUC} & COVID-19 & 0.551 & 0.553 & 0.555 & 0.544 & 0.541 & \textbf{0.564} \\
\hline
\multirow{1}{*}{F1} & COVID-19 & 0.663 & 0.540 & 0.663 & 0.660 & \textbf{0.665} & 0.662 \\
\hline
\multirow{1}{*}{ACC} & COVID-19 & 0.524 & \textbf{0.540} & 0.529 & 0.531 & 0.527 & 0.530 \\
\hline
\end{tabular}
\label{tab9}
\end{table}

\clearpage

\end{document}